\documentclass[titlepage]{csetr}

\usepackage{graphicx}
\usepackage{amsmath,amsfonts,amsthm,amssymb}
\usepackage{setspace}
\usepackage{url}
\usepackage{fancyhdr}
\usepackage{lastpage}
\usepackage{multicol}
\usepackage{pdfpages}
\usepackage{color}
\usepackage{xcolor}
\usepackage[normalem]{ulem}
\usepackage{booktabs}
\usepackage{multirow}
\usepackage{algorithm}
\usepackage[noend]{algpseudocode}

\renewcommand{\and}{\hspace{.5cm}}

\trnumber{201517}
\trdate{November 2015}

\title{%
  On Improving Informativity and Grammaticality for Multi-Sentence Compression
}
\author{%
  Elahe Shafiei \and %
  Mohammad Ebrahimi \and %
  Raymond Wong\\[1em]
  Fang Chen\\[2em]
  University of New South Wales, Australia \\%
  ATP Laboratory, National ICT, Sydney, Australia \\%
  \email{\{elahehs,mohammade,wong,fang\}@cse.unsw.edu.au}\\[3cm]
}

\date{}
\begin{document}
\maketitle

\begin{abstract}
  Multi Sentence Compression (MSC) is of great value to many real world applications, such as guided microblog summarization, opinion summarization and newswire summarization. Recently, word graph-based approaches have been proposed and become popular in MSC. Their key assumption is that redundancy among a set of related sentences provides a reliable way to generate informative and grammatical sentences. In this paper, we propose an effective approach to enhance the word graph-based MSC and tackle the issue that most of the state-of-the-art MSC approaches are confronted with: i.e., improving both informativity and grammaticality at the same time. Our approach consists of three main components: (1) a merging method based on Multiword Expressions (MWE); (2) a mapping strategy based on synonymy between words; (3) a re-ranking step to identify the best compression candidates generated using a POS-based language model (POS-LM). We demonstrate the effectiveness of this novel approach using a dataset made of clusters of English newswire sentences. The observed improvements on informativity and grammaticality of the generated compressions show that our approach is superior to state-of-the-art MSC methods.

\end{abstract}

\section{Introduction}

Multi-Sentence Compression (MSC) refers to the method of mapping a collection of related sentences to a sentence shorter than the average length of the input sentences, while retaining the most important information that conveys the gist of the content, and still remain grammatically correct \cite{jing2000sentence,boudin2013keyphrase}. MSC is one of the challenging tasks in natural language processing that has recently attracted increasing interest \cite{boudin2013keyphrase}. This is mostly because of its potential use in various applications such as guided microblog summarization, opinion summarization, newswire summarization, text simplification for mobile devices and so on. A standard way to generate summaries usually consists of the following steps: ranking sentences by their importance, clustering them by similarity, and selecting a sentence from the top ranked clusters \cite{wang2008multi}.

Traditionally, most of the MSC approaches rely on syntactic parsers, e.g. \cite{filippova2008sentence,elsner2011learning}. As an alternative, some recent works in this field \cite{filippova2010multi,boudin2013keyphrase} are based on word graphs, which only require a Part-Of-Speech (POS) tagger and a list of stopwords. These approaches simply rely on the words of the sentences and efficient dynamic programming. They take advantage of the redundancy among a set of related sentences to generate informative and grammatical sentences. 

Although the proposed approach in \cite{filippova2010multi} introduces an elegant word graph to MSC, approximately half of their generated sentences are missing important information about the set of related sentences \cite{boudin2013keyphrase}. Afterwards, Boudin and Morin (2013) enhanced their work and produced more informative sentences by maximizing the range of topics they cover. However, they confirmed that grammaticality scores are decreased, since their re-ranking algorithm produces longer compressions to ameliorate informativity. Therefore, grammaticality might be sacrificed while enhancing informativity and vice versa. 

\begin{figure}[h!]
	\centering
	\includegraphics[scale=0.7]{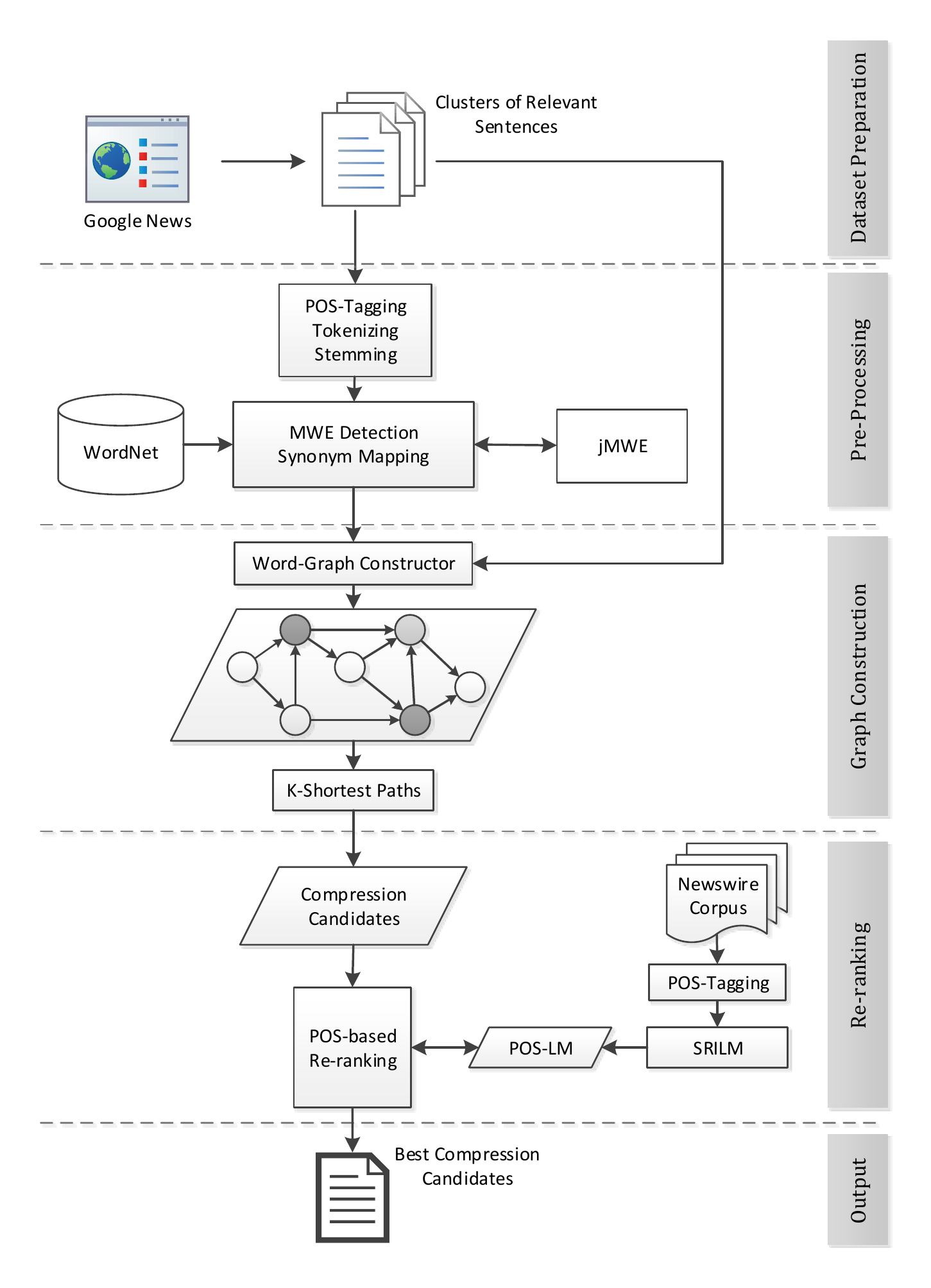}
	\caption{Overview of the proposed approach}
	\label{fig:overview}
	\vspace{0.5cm}
\end{figure}

In this paper, we are motivated to tackle the main difficulty of the above mentioned MSC approaches which is to simultaneously improve both informativity and grammaticality of the compressed sentences. To this end, we propose a novel enhanced word graph-based MSC approach by employing significant merging, mapping and re-ranking steps that favor more informative and grammatical compressions. The contributions of the proposed method can be summarized as follows: (1) we exploit Multiword Expressions (MWE) from the given sentences and merge their words, constructing each MWE into a specific node in the word graph to reduce the ambiguity of mapping, so that well-organized and more informative compressions can be produced; (2) we take advantage of the concept of synonymy in two ways: firstly, we replace a merged MWE with its \textit{one}-word synonym if available, and secondly, we use the synonyms of an upcoming single word to find the most proper nodes for mapping; (3) we employ a 7-gram POS-based language model (POS-LM) to re-rank the \textit{k}-shortest obtained paths, and produce well-structured and more grammatical compressions.  To our knowledge, this paper presents the first attempt to use MWEs, synonymy and POS-LM to improve the quality of word graph-based MSC. Extensive experiments on the released standard dataset demonstrate the effectiveness of our proposed approach. Figure \ref{fig:overview} also depicts the overview of this approach.

The rest of this paper is organized as follows. Section \ref{relatedw} summarizes the related work. Section \ref{proposeda} presents our proposed approach. The data preparation process for evaluating our method is demonstrated in Section \ref{datap}, and Section \ref{experiments} reports the evaluation metrics and the performed experiments. Finally, Section \ref{conclusions} concludes the paper.

\section{Related Work}\label{relatedw}

\subsection{Multi-Sentence Compression}\label{MSC}

State-of-the-art approaches in the field of MSC are generally divided into supervised \cite{mcdonald2006discriminative,galley2007lexicalized} and unsupervised groups \cite{clarke2007modelling}. MSC methods traditionally use a syntactic parser to generate grammatical compressions, and fall into two categories (based on their implementations): (1) \textit{tree-based} approaches, which create a compressed sentence by making edits to the syntactic tree of the original sentence \cite{mcdonald2006discriminative,galley2007lexicalized,filippova2008sentence,elsner2011learning}; (2) \textit{sentence-based} approaches, which generates strings directly \cite{clarke2007modelling}. 

As an alternative, word graph-based approaches that only require a POS tagger have recently been used in different tasks, such as guided microblog summarization \cite{sharifi2010experiments}, opinion summarization \cite{ganesan2010opinosis} and newswire summarization \cite{filippova2010multi,boudin2013keyphrase,tzouridis2014learning}. In these approaches, a directed word graph is constructed in which nodes represent words while edges between two nodes represent adjacency relations between words in a sentence. Hence, the task of sentence compression is performed by finding the k-shortest paths in the word graph. In particular, our work is applied to newswire summarization. In this field, Filippova (2010) has introduced an elegant word graph-based MSC approach that relies on the redundancy among the set of related sentences. However, some important information are missed from 48\% to 60\% of the generated sentences in their approach \cite{boudin2013keyphrase}. Thus, Boudin and Morin (2013) proposed an additional re-ranking scheme to identify summarizations that contain key phrases. However, they mentioned that grammaticality is sacrificed to improve informativity in their work.

In our proposed approach, we utilize MWEs and synonym words in sentences to significantly enhance the traditional word graph, and improve informativity. Then, we re-rank the generated compression candidates with a 7-gram POS-LM that captures the syntactic information, and strengthens the compressed sentences in terms of grammaticality. 
 
\subsection{Multiword Expressions}\label{MWE}

An MWE is a combination of words with lexical, syntactic or semantic idiosyncrasy \cite{sag2002multiword,baldwin2010multiword}. It is estimated that the number of MWEs in the lexicon of a native speaker of a language has the same order of magnitude as the number of single words \cite{jackendoff1997architecture}. Hence, explicit identification of MWEs has been shown to be useful in various NLP applications. Components of an MWE can be treated as a single unit to improve the effectiveness of re-ranking steps in IR systems \cite{acosta2011identification}. In this paper, we identify MWEs, merge their components, and replace them with their available \textit{one}-word synonyms, if applicable. These strategies help to construct an improved word graph and enhance the informativity of the compression candidates.

\subsection{POS-based Language Model (POS-LM)}\label{POS-LM}

A language model assigns a probability to a sequence of \emph{m} words $P(w_1, ..., w_m)$ by means of a probability distribution. Language models are an essential element of natural language processing, in tasks ranging from spell-checking to machine translation. Given the increasing need to ensure grammatical sentences in different applications, POS-LM comes into play as a remedy. POS-LM describes the probability of a sequence of \emph{m} POS tags $P(t_1, ..., t_m)$.  POS-LMs are traditionally used for speech recognition problems \cite{heeman1998pos} and statistical machine translation systems \cite{koehn2008towards,monz2011statistical,popovic2012morpheme} to capture syntactic information. In this paper, we benefit from POS-LMs to capture the syntactic information of sentences and improve the grammaticality of compression candidates.

\section{Proposed Approach}\label{proposeda}

\subsection{Word Graph Construction for MSC}\label{wgMSC}

Consider a set of related sentences $S=\{s_1,s_2,...,s_n\}$, a traditional word graph is constructed by iteratively adding sentences to it. This directed graph is an ordered pair $G=(V,E)$ comprising of a set of vertices or words together with a set of directed edges which shows the adjacency between corresponding nodes \cite{filippova2010multi,boudin2013keyphrase}. The graph is firstly constructed by the first sentence and displays words in a sentence as a sequence of connected nodes. The first node is the start node and the last one is the end node. Words are added to the graph in three steps of the following order: (1) non-stopwords for which no candidate exists in the graph; or for which an unambiguous mapping is possible (i.e. there is only one node
in the graph that refer to the same word/POS pair); (2) non-stopwords for which there are either several possible candidates in the graph; or for which they occur more than once in the sentence; (3) stopwords. For the last group, same as Boudin and Morin (2013), we use the stopword list included in nltk\footnote{http://nltk.org/} extended with temporal nouns such as \lq yesterday\rq, \lq Friday\rq, and etc..

All MSC approaches aim at producing condensed sentences that inherit the most important information from the original content while remains syntactically correct. However, gaining these goals at the same time remains still difficult. As a remedy, we believe that a better resolution to construct an improved word graph can be obtained by using more sophisticated pre-processing and re-ranking steps. Thus, we focus on the notions of synonymy, MWE and POS-LM re-ranking, which dramatically raise the informativity and grammaticality of compression candidates. In the following, we describe the details of our proposed approach:

\subsection{Merging and Mapping Strategies}\label{mergmap}

Like many NLP applications, MSC will benefit from the identification of MWEs and the concept of synonymy; and even more so when lexical diversity arises in a collection of sentences. For example, consider a sentence that includes an MWE (\textit{kick the bucket}): \textit{It would be a sad thing to \underline{kick the bucket} without having been to Alaska}. To benefit from this MWE that has 3 components/words, we propose the merging strategy below:

Firstly, after tokenizing the sentence and stemming the words, we detect the MWE and its tuple POS with an MWE detector. This step has the advantage of reducing the ambiguity of mapping upcoming words onto the existing words with the same appearance in the graph. For example, the word \textit{kick} above has a different meaning and POS (as an MWE component) from the identical appearance word \textit{kick} in isolation (in another sentence say “they \textit{kick} open the door and entered the room.”). So, MWE identification can keep us from mapping these two \textit{kick} together and retain the important meaning of the content. To detect MWEs, we use the jMWE toolkit \cite{kulkarni2011jmwe}, which is a Java-based library for constructing and testing MWE token detectors. 

\begin{figure}[h!]
	\centering
	\includegraphics[scale=0.7]{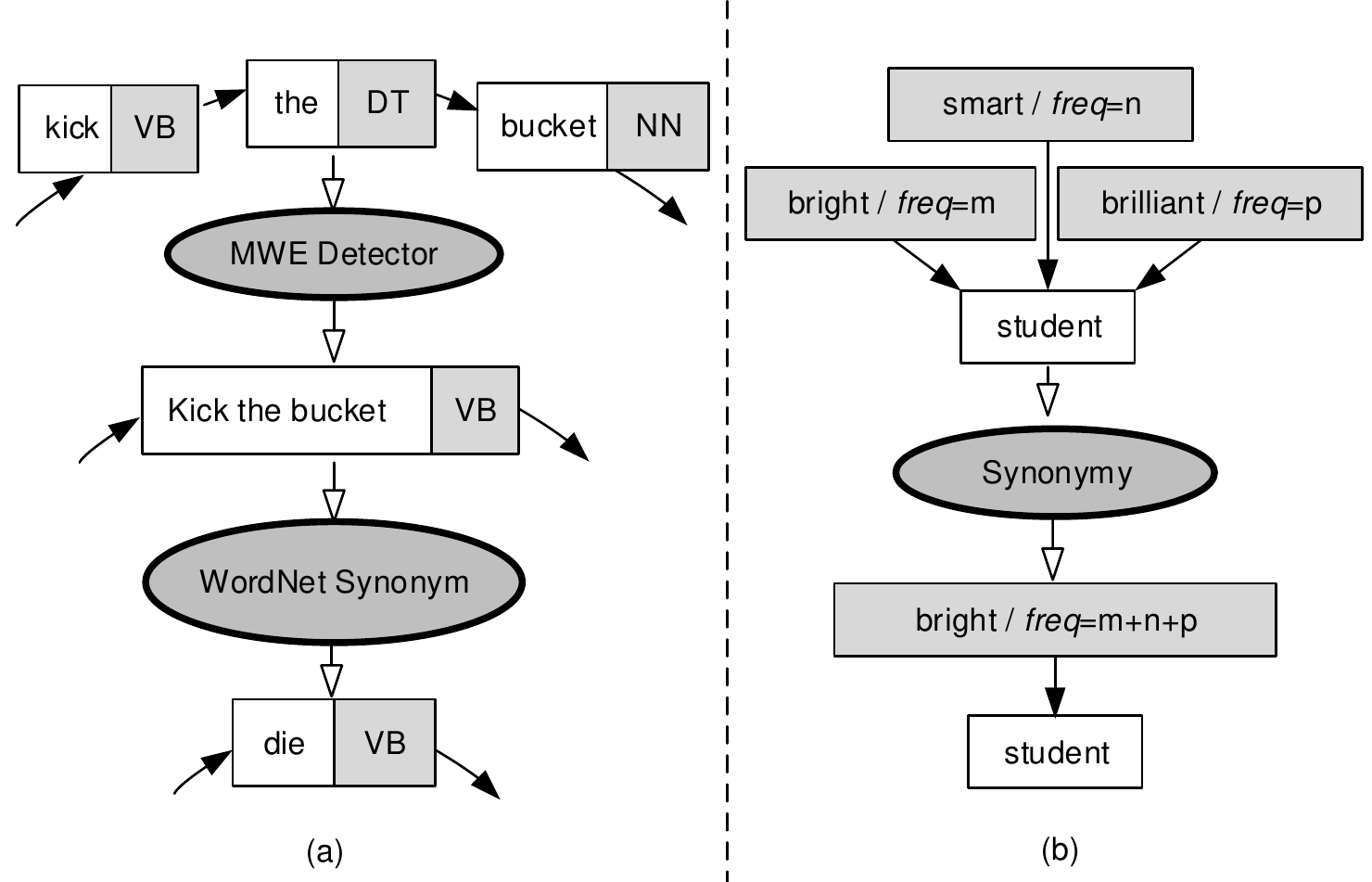}
	\caption{(a) Example of MWE merging and mapping, (b) Example of Synonym mapping}
	\label{fig:mergmap}
	\vspace{0.5cm}
\end{figure}

Secondly, we use version 3.0 of WordNet \cite{miller1995wordnet} to obtain its available \textit{one}-word synonym with an appropriate POS and replace the \textit{n}-words MWE with a shorter synonym word. WordNet groups all synonyms into a SynSet - a synonym set. We only consider the most frequent \textit{one}-word synonym in the WordNet that also appears in the other relevant sentences. If other relevant sentences contain none of the \textit{one}-word synonyms, the most frequent one is selected directly from the WordNet to help condense the sentence. Three native speakers were asked to investigate all the synonym mappings performed in our approach, and specify whether each mapped synonym reflects the meaning of the original word in the sentence or not. Based on this evaluation, the average rate of correct synonym mappings is 88.21\%. In case that no appropriate synonym is found for MWE, the merged MWE itself was used as a back-off. This can reduce the number of graph nodes and, consequently, the ambiguity for further false mappings of MWE components in the word graph. These steps are briefly depicted in Figure \ref{fig:mergmap} (a).

Furthermore, we use the concept of synonymy for mapping upcoming single words. For example, consider 3 different sentences containing words \textit{bright}, \textit{smart} and \textit{brilliant}, which are synonyms of each other. Assume each sentence contains one of these synonyms respectively. Without an appropriate mapping based on a notion of synonymy, these 3 nodes will be added to the word graph as separate nodes. With our approach, the word graph in this example is constructed with a single node containing a word as a representative of its synonyms from the other sentences. The weight of the obtained node is computed by summing the frequency scores from the other nodes as shown in Figure \ref{fig:mergmap} (b) for each pair of word/POS. The main purpose of this modification is three fold: (i) the ambiguity of mapping nodes is reduced; (ii) the number of total possible paths (compression candidates) is decreased; and (iii) the weight of frequent similar words with different appearances in the content is better reflected by the notion of synonymy. 

In the following example, we will demonstrate how we use the pre-processing strategies to produce refined sentences, and generate an improved word graph. Among the underlined words, MWEs are put into bracket, and synonyms are identified by the same superscript notations.\\[0.1em]

\noindent(1) \uwave{Teenage}\textsuperscript a boys are more \uwave{interested}\textsuperscript b in \uwave{[junk food]}\textsuperscript c marketing and \uwave{consume}\textsuperscript d more \uwave{[fast food]}\textsuperscript c than girls.

\noindent(2) \uwave{[Junk food]}\textsuperscript c marketers find \uwave{young}\textsuperscript a boys more \uwave{fascinated}\textsuperscript b than girls, a survey \uwave{released}\textsuperscript e by the Cancer Council shows.

\noindent(3) \uwave{Adolescent}\textsuperscript a boys \uwave{[use up]}\textsuperscript d more \uwave{[fast food]}\textsuperscript c than girls, \uwave{[according to]} a new survey.

\noindent(4) The survey, \uwave{published}\textsuperscript e by the Cancer Council, observed \uwave{teenage}\textsuperscript a boys were regular consumers of \uwave{[junk food]}\textsuperscript c.\\[0.1em]

\begin{figure}[h!]
	\centering
	\includegraphics[scale=0.8]{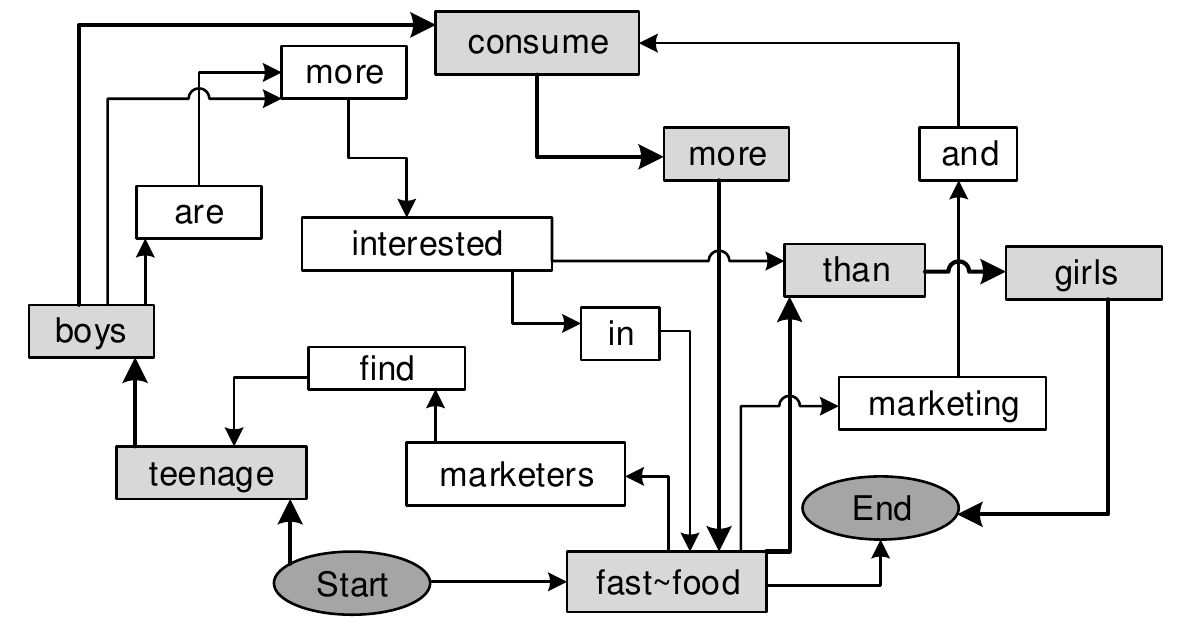}
	\caption{The generated word graph and a compression path}
	\label{fig:wgraph}
	\vspace{0.5cm}
\end{figure}

The word graph constructed for the above sentences are partially shown in Figure \ref{fig:wgraph}. Some nodes, edge weights and punctuations are omitted from the graph for more clarity.

Where mapping in the graph is ambiguous (i.e. there are two or more nodes in the graph that refer to the same word/POS pair), we follow the instruction stated by Filippova (2010): the immediate context (the preceding and following words in the sentence, and the neighboring nodes in the graph) or the frequency (i.e. the node which has words mapped to it) is used to select the best candidate node. A new node is created only if there is no suitable candidate to be mapped to, in the graph.

In Filippova (2010), edge weights are calculated using the weighting function defined in Equation \ref{eq:weigh1} in which $w^{'}(e_{i,j})$ is given by Equation \ref{eq:weigh2}.

\begin{equation}
w(e_{i,j})=\frac{w^{'}(e_{i,j})}{freq(i) \times freq(j)}
\label{eq:weigh1}
\end{equation}

\begin{equation}
w^{'}(e_{i,j})=\frac{freq(i) + freq(j)}{\sum_{s \in S}{diff(s,i,j)^{-1}}}
\label{eq:weigh2}
\end{equation}

\noindent where $freq(i)$ is the number of words mapped to the node \emph{i}. The function $diff(s,i,j)$ refers to the distance between the offset positions of words \emph{i} and \emph{j} in sentence \emph{s}.\\[0.1em]

\makeatletter
\def\BState{\State\hskip-\ALG@thistlm}
\makeatother
\begin{algorithm}[h!]
	
	\caption{Proposed MSC Word Graph}\label{alg:MSCgraph}
	\begin{algorithmic}[1]
		\State $\textbf{Input: }\text{A cluster of relevant sentences: } S=\{s_i\}^n_{i=1}$
		\State $\textbf{Output: }G=(V, E)$ 
		\State \textbf{for} $i=1$ to $n$ \textbf{do}
		\State \indent $\textit{t} \gets \textit{Tokenize}({s_i})$ 
		\State \indent $\textit{st} \gets \textit{Stemming}(t)$ 
		\State \indent $\textit{MWE-comp} \gets \textit{MWE-Detection}(t, st)$ 
		\State \indent $\textit{MWE-list} \gets \textit{Merge-MWE}(MWE$-$comp)$ 
		\State \indent $\textit{sentSize} \gets \textit{SizeOf}(t)$ 
		\State \indent \textbf{for} $j=1$ to $sentSize$ \textbf{do}
		\State \indent \indent $\textit{LABEL} \gets t_j$ 
		\State \indent \indent $\textit{SID} \gets i$ 
		\State \indent \indent $\textit{PID} \gets j$ 
		\State \indent \indent $\textit{SameN} \gets getSameNodes(G, LABEL)$
		\State \indent \indent \textbf{if} $sizeOf(SameN) \geq 1$ \textbf{then}
		\State \indent \indent \indent ${v_j} \gets \textit{getBestSame}(SameN)$ 
		\State \indent \indent \indent $\textit{mapList}_{v_j} \gets \textit{mapList}_{v_j} \cup (SID, PID)$ 
		\State \indent \indent \textbf{else} 
		\State \indent \indent \indent $\textit{SynN} \gets \textit{getSynonymNodes}(G, LABEL)$ 
		\State \indent \indent \indent \textbf{if} $sizeOf(SynN) \geq 1$ \textbf{then}
		\State \indent \indent \indent \indent ${v_j} \gets \textit{getBestSyn}(SynN)$
		\State \indent \indent \indent \indent $\textit{mapList}_{v_j} \gets \textit{mapList}_{v_j} \cup (SID, PID)$ 
		\State \indent \indent \indent \textbf{esle if} $t_j \in$ \textit{MWE-list} \textbf{then}
		\State \indent \indent \indent \indent $\textit{WNSyn} \gets \textit{getBestWNSyn}(LABEL)$
		\State \indent \indent \indent \indent ${v_j} \gets \textit{creatNewNode}(G,\textit{WNSyn})$
		\State \indent \indent \indent \indent $\textit{mapList}_{v_j} \gets (SID, PID)$ 
		\State \indent \indent \indent \textbf{esle} 
		\State \indent \indent \indent \indent ${v_j} \gets \textit{creatNewNode}(G,LABEL)$
		\State \indent \indent \indent \indent $\textit{mapList}_{v_j} \gets (SID, PID)$ 
		\State \indent \indent \indent \textbf{end if} 
		\State \indent \indent \textbf{end if} 
		\State \indent \indent \textbf{if} \textit{not existEdge}$(G, {v_{j-1}} \to {v_j})$ \textbf{then}
		\State \indent \indent \indent \textit{addEdge}$({v_{j-1}} \to {v_j}, G)$ 
		\State \indent \indent \textbf{end if}  
		\State \indent \textbf{end for} 
		\State \textbf{end for}  
	\end{algorithmic}
\end{algorithm}

Algorithm \ref{alg:MSCgraph} presents the steps to build our proposed MSC word graph, G(V, E). We start with a cluster of relevant sentences from a set of input newswire clusters. Each cluster is denoted as  
$S=\{s_i\}^n_{i=1}$ where each $s_i$ is a sentence containing POS annotations. \textit{Line 4-5:} Each $s_i \in S$ is split into a set of tokens, where each token, $t_j$ consists of a word and its corresponding POS annotation (e.g.\emph{"boys:NN"}). The tokens are also stemmed into a set of stemmed words, \emph{st}. \textit{Line 6-7:} For each sentence, MWE components, i.e., \emph{MWE-comp}, are detected using the set of tokens \emph{t} and stems \emph{st}. Then, these MWE components are merged in each sentence, and kept in a list of \emph{MWE-list}. \textit{Line 10-12:} Each unique $t_j$ will form a node $v_j$ in the MSC graph, with $t_j$ being the label. Since we only have one node per unique token, each node keeps track of all sentences that include its token. So, each node keeps a list of \emph{sentence identifier}, (SID) along with the \emph{position of token} in that sentence, (PID). Each node including a single word or a merged MWE will thus carry a \emph{mapping list} (mapList) which is a list of \{SID:PID\} pairs representing the node's membership in a sentence.

 \textit{Line 13-16:} For mapping the token $t_j$, we first explore the graph to find the same node (i.e. node that refers to the same word/POS pair as $t_j$). If two or more same nodes are found, considering the aforementioned ambiguous mapping criteria in Section \ref{mergmap}, the best candidate node is selected for mapping. Then the pair of (SID:PID) of $t_j$ will be added to the mapping list of the selected node, i.e., $mapList_{v_j}$. \textit{Line 18-21:} If no same node exists in the graph, then we look for the best synonym node in the graph (i.e. find the most frequent synonym among the WordNet synsets that was earlier added to the graph.). Again, the mapping list of the selected node, $mapList_{v_j}$ will be updated to include the pair of (SID:PID) of $t_j$. \textit{Line 22-28:} If none of the above conditions are satisfied, it is time to create a new node in the graph. However as explained in Section \ref{mergmap}, when $t_j$ is MWE, we extract the best WordNet \textit{one}-word synonym, and replace the \textit{n}-word MWE with this shorter synonym word. So, a shorter content node will be added to the graph. \textit{Line 31-33:} the original structure of a sentence is reordered with the use of directed edges.

A heuristic algorithm is then used to find the \textit{k}-shortest paths from start to end node in the graph. Throughout our experiments, the appropriate value for \emph{k} is 150. By re-ranking this number of shortest paths, most of the potentially good candidates are kept and a decline in performance is prevented. Paths shorter than eight words or do not contain a verb are filtered before re-ranking. The remaining paths are re-ranked and the path that has the lightest average edge weight is eventually considered as the best compression. Next, an accurate re-ranking approach to identify the most informative grammatical compression candidate is described.
       
\subsection{Re-ranking Strategy (POS-LM)}\label{rerank}

Boudin and Morin (2013) have recently utilized TextRank (Mihalcea and Tarau 2004) to re-rank the compression candidates. In their approach, a word recommends other co-occurring words, and the strength of the recommendation is recursively computed based on the importance of the words making the recommendation. The score of a keyphrase \emph{k} is computed by summing the salience of the words it contains, normalized with its $length+1$ to favor longer \emph{n}-grams according to Equation \ref{eq:scoreK}.

\begin{equation}
score(k)=\frac{\sum_{w \in k}{TextRank(w)}}{length(k)+1}
\label{eq:scoreK}
\end{equation}\\[0.1em] 

\noindent Finally, the paths are re-ranked and the score of a compression candidate \emph{c} is given by Equation \ref{eq:scoreC}.

\begin{equation}
	score(c)=\frac{\sum_{{i,j} \in path(c)}{w(e_{i,j})}}{length(c) \times \sum_{k \in c}{score(k)}}
\label{eq:scoreC}
\end{equation}\\[0.1em]

In our re-ranking step, we benefit from the fact that POS tags capture the syntactic roles of words in a sentence. We use a POS-LM to assign a grammaticality score to each generated compression. Our hypothesis is that POS-LM helps in identifying the most grammatical sentence among the \emph{k}-most informative compressions. This strategy shall improve the grammaticality of MSC, even when the grammatical structures of the input sentences are completely different. Word-based language models estimate the probability of a string of \emph{m} words by Equation \ref{eq:lm}, and POS-LMs estimate the probability of string of \emph{m} POS tags by Equation \ref{eq:poslm} \cite{monz2011statistical}. 

\begin{equation}
p(w_1^m) \propto \prod_{i-1}^{m} p(w_i|w_{i-n+1}^{i-1})
\label{eq:lm}
\end{equation}

\begin{equation}
p(t_1^m) \propto \prod_{i-1}^{m} p(t_i|t_{i-n+1}^{i-1})
\label{eq:poslm}
\end{equation}

\noindent where, \emph{n} is the order of the language model, and \emph{w/t} refers to the sub-sequence of words/tags from position \emph{i} to \emph{j}.\\[0.5em]

To build a POS-LM, we use the SRILM toolkit with modified Kneser-Ney smoothing \cite{stolcke2002srilm}, and train the language model on our POS annotated corpus. SRILM collects \emph{n}-gram statistics from all \emph{n}-grams occurring in a corpus to build a single global language model. To train our POS-LM, we need a POS-annotated corpus. In this regard, we make use of the Stanford POS tagger \cite{toutanova2003feature} to annotate the AFE sections of LDC’s Gigaword corpus (LDC2003T05) as a large newswire corpus ($\sim$170 M-words). Then, we remove all words from the pairs of words/POS in the POS annotated corpus. 

Although the vocabulary of a POS-LM, which is usually ranging between 40 and 100 tags, is much smaller than the vocabulary of a word-based language model, there is still a chance in some cases of unseen events. Since modified Kneser-Ney discounting appears to be the most efficient method in a systematic description and comparison of the usual smoothing methods \cite{goodman2001bit}, we use this type of smoothing to help our language model. 

The compression candidates also need to be annotated with POS tags. So, the score of each compression is estimated by the language model, based on its sequence of POS tags. Since factors like POS tags, are less sparse than surface forms, it is possible to create a higher order language models for these factors. This may encourage more syntactically correct output \cite{koehn2007moses}. Thus, in our approach we use 7-gram language modeling based on part of speech tagging to re-rank the \emph{k}-best compressions generated by the word graph. 

To re-rank the obtained paths, our POS-LM gives the perplexity score ($score_{LM}$) which is the geometric average of \emph{1/probability} of each sentence, normalized by the number of words. So, $score_{LM}$ for each sequence of POS in the \emph{k}-best compressions is computed by Equation \ref{eq:scoreLM}.

\begin{equation}
score_{LM}(c)=10^{\frac{\log_{}{prob(c)}}{\#word}}
\label{eq:scoreLM}
\end{equation}

\noindent where $prob(c)$ is the probability of compression \emph{(C)} including $\#Word$ number of words, computed by the 7-gram POS-LM.

As the estimated scores for each cluster of sentences fall
into different ranges, we make use of unity-based normalization to bring the values of $score(c)$ in Equation 4, and the $score_{LM}$ into the range [0, 1]. The score of each compression is finally given by Equation \ref{eq:final} 

\begin{equation}
score_{final}(c)=\mu \times score(c) + (1-\mu) \times score_{LM}(c)
\label{eq:final}
\end{equation}

\noindent in which the scaling factor $\mu$ in our experiments has been set to 0.4, so as to reach the best re-ranking results.\\[0.1em]

To better understand how POS-LM is used, consider the sentences below, which have the same scores for informativity but are added into our re-ranking contest to be investigated based on their grammaticality. The corresponding POS sequences of these sentences are given to the trained language model to clarify which one is more grammatical.\\[0.1em]

\begin{center}
	\begin{tabular}{ c c c c c c c }
		(1) & Boys & $\textcolor{darkgray}{more \enspace \enspace consume}$ & fast & food & than & girls. \\ 
		  & NNS & $\textcolor{darkgray}{\underbrace{RBR \quad \enspace \enspace VBP}}$ & JJ & NN & IN & NNS \\  
		  &   & \textcolor{gray}{Wrong Pattern} &   &   &   & \\ 
		  &   &   &   &   &   & \\ 
		(2) & Boys & $consume \enspace \enspace more$ & fast & food & than & girls. \\ 
		  & NNS & $VBP \quad \enspace \enspace JJR$ & JJ & NN & IN & NNS \\
	\end{tabular}
\end{center}
      
As expected, the winner of this contest is the second POS sequence, which has a better grammatical structure and gets a higher probability score from the POS-LM.

\section{Data Preparation}\label{datap}

Many attempts have been made to release various kinds of datasets and evaluation corpora for sentence compression and automatic summarization, such as the one introduced in \cite{clarke2006models}. However, to our knowledge, there is no dataset available to evaluate MSC in an automatic way \cite{boudin2013keyphrase}. Since the prepared dataset in Boudin and Morin (2013) is also in French, we have followed the below instructions to construct a Standard English newswire dataset: 

We have collected news articles in clusters on the Australian\footnote{http://news.google.com.au/} and U.S.\footnote{http://news.google.com/} edition of Google News over a period of five months (January 2015 - May 2015). Clusters composed of at least 15 news articles about one single news event, were manually extracted from different categories (i.e. Top Stories, World, Business, Technology, Entertainment, Science, Health, etc.). Leading sentences in news articles are known to provide a good summary of the article content and are used as a baseline in summarization \cite{dang2005overview}. Hence, to obtain the sets of related sentences, we have extracted the first sentences from the articles in the cluster and removed duplicates. 

The released dataset contains 568 sentences spread over 46 clusters (each is related to one single news event). The average number of sentences within each cluster is 12, with a minimum of 7 and a maximum of 24. Three native English speakers were also asked to meticulously read the sentences provided in the clusters, extract the most salient facts, summarize the set of sentences, and generate three reference summaries for each cluster with as less new vocabularies as possible.  

In practice, along with the clusters of sentences with similar lexical and grammatical structures (we refer to these clusters as \textit{normal}), it is likely to have clusters of content-relevant sentences, but with different (non-redundant) appearances and grammatical structures (we consider these clusters as \textit{diverse}). In fact, the denser a word graph is, the more edges interconnect with vertices and hence more paths pass through the same vertices. This results in low lexical and syntactical diversity, and vice versa \cite{tzouridis2014learning}. The density of a word graph generated by sentences of a cluster $G=(V,E)$ is given by Equation \ref{eq:density}.

\begin{equation}
Density=\frac{|E|}{|V|(|V|-1)}
\label{eq:density}
\end{equation}

\noindent Thereupon, we have also identified 15 \textit{diverse} clusters among the 46 clusters to demonstrate the effect of our approach on the normal and diverse groups. Table \ref{tbl:dataset} lists the properties of the evaluation dataset.

\begin{table}[h!]
\begin{center}
	\begin{tabular}{|| c c ||} 
		\hline
		total \#clusters & 46 \\ 
		\hline\hline
		\#normal clusters & 31 \\
		\hline\hline
		\#diverse clusters & 15 \\
		\hline\hline
		total \#sentences & 568 \\
		\hline\hline
		avg \#sentences/cluster & 12 \\ 
		\hline\hline
		min \#sentences/cluster & 7 \\
		\hline\hline
		max \#sentences/cluster & 24 \\[1ex] 
		\hline
	\end{tabular}
	\caption{Information about the constructed dataset}
	\label{tbl:dataset}
\end{center}
\end{table}

\section{Experiments}\label{experiments}

\subsection{Evaluation Metrics}\label{metric}

We evaluate the proposed method over our constructed dataset (\textit{normal} and \textit{diverse} clusters) using automatic and the manual evaluations. The quality of the generated compressions was assessed automatically through version 2.0 \footnote{http://kavita-ganesan.com/content/rouge-2.0} of \textsc{Rouge} \cite{lin2004rouge} and the version 13a \footnote{ftp://jaguar.ncsl.nist.gov/mt/resources/mteval-v13a.pl} of \textsc{Bleu}  \cite{papineni2002bleu}. These sets of metrics are typically used for evaluating automatic summarization and machine translation. They compare an automatically produced summary against a reference or a set of human-produced summaries. 

For the manual investigation of the quality of the generated compressions, three native English speakers were asked to rate the grammaticality and informativity of the compressions based on the points scale defined in Filippova (2010). \textit{Grammaticality}: (i) if the compression is grammatically perfect $\rightarrow$ \textit{point 2}; (ii) if the compression requires some minor editing $\rightarrow$ \textit{point 1}; (iii) if the compression is ungrammatical $\rightarrow$ \textit{point 0}. The lack of capitalization is ignored by the raters. \textit{Informativity}: (i) if the compression conveys the gist of the content and is mostly similar to the human-produced summary $\rightarrow$ \textit{point 2}; (ii) if the compression misses some important information $\rightarrow$ \textit{point 1}; (iii) if the compression contains none of the important contents $\rightarrow$ \textit{point 0} (Table \ref{tbl:agreement}). 

The \textit{k} value for the agreement between raters falls into range (0.4 $\sim$ 0.6) through Kappa's evaluation metrics, which indicates that the strength of this agreement is moderate \cite{artstein2008inter}.

\begin{table}[h!]
	\begin{center}
\begin{tabular}{@{}llllll@{}}
	\toprule
	\midrule
	\textbf{ Feature} & \textbf{State of the Compression} &\multicolumn{3}{c}{\textbf{Point}} \\
	\midrule
	&& 2 & 1 & 0 & \\
	\cmidrule{3-6} 
	\multirow{3}{*}{\textbf{ Grammaticality \Bigg\{}} & grammatically perfect & $\surd$\\
	& requires some minor editing & &$\surd$ \\
	& ungrammatical & & &$\surd$\\
	\midrule
	\multirow{3}{*}{\textbf{ Informativity \Bigg\{}} & conveys the gist of the content & $\surd$\\
	& misses some important information & &$\surd$ \\
	& contains none of the important contents  & & &$\surd$\\
	\midrule
	\bottomrule
\end{tabular}
	\caption{Points scale defined in the agreement between raters}
	\label{tbl:agreement}
\end{center}
\end{table}

\subsection{Experiment Results}\label{experiment}

Two existing approaches, i.e., Filippova (2010) and Boudin and Morin (2013) are used as Baseline1 and Baseline2 respectively, for comparison purposes in our experiments. To better understand the behavior of our system, we examined our test dataset, and made the following observations. For the manual evaluation (Table \ref{tbl:avg_manual_scores}), we observed a significant improvement in the average grammaticality and informativity scores along with the compression ratio (CompR) over the normal and diverse clusters. The informativity of Baseline1 is adversely influenced by missing important information about the set of related sentences \cite{boudin2013keyphrase}. However Baseline2 enhanced the informativity, the grammaticality scores are decreased due to the outputs of longer compressions. In our approach, the remarkable improvement in the grammaticality scores is due to the adding of the syntactic-based re-ranking step. Using this re-ranking method, the most grammatical sentences are picked among the \textit{k}-best compression candidates. Furthermore, merging MWEs, replacing them with their available \textit{one}-word synonyms and mapping words using synonymy all enhance the informativity scores, and help to generate a denser word graph instead of a sparse one. Given that, the value of the compression ratio ($\sim$48\%) is better than the best obtained compression ratio on these two baselines (50\%). 

\begin{table}[h!]
	\begin{center}	
	\begin{tabular}{l*{6}{c}}
		\toprule
		\midrule
		\textbf{Method} &\multicolumn{2}{c}{\textbf{Normal}}&\multicolumn{2}{c}{\textbf{Diverse}}&  \textbf{CompR}\\
		\cmidrule(lr){2-5}
		 & \textbf{Info.} & \textbf{Gram.} & \textbf{Info.} & \textbf{Gram.} \\
		\midrule
		Baseline1 & 1.44 & 1.67 & 1.17 & 1.19 & 50\% \\
		Baseline2 & 1.68 & 1.60 & 1.30 & 1.12 & 58\% \\
		Proposed & 1.68 & 1.68 & 1.36 & 1.47 & 48\% \\
		\midrule
		\bottomrule
	\end{tabular}
	\caption{Average scores over normal and diverse clusters separately given by the raters; along with the estimated compression rate}
	\label{tbl:avg_manual_scores}
\end{center}
\end{table}

The average performance of the baseline methods and the proposed approach over the normal and diverse clusters in terms of \textsc{Rouge} and \textsc{Bleu} scores are also shown in Table \ref{tbl:avg_auto_scores}. \textsc{Rouge} measures the concordance of candidate and reference summaries by determining \textit{n}-gram, word sequence, and word pair matches. We used \textsc{Rouge} F-measure for unigram, bigrams, and \textsc{su4} (skip-bigram with maximum gap length 4) to evaluate the compression candidates. The \textsc{Bleu} metric computes the scores for individual sentences; then averages these scores over the whole corpus for a final score. We used \textsc{Bleu} for 4-grams to evaluate the results.

\begin{table}[h!]
	\begin{center}	
		\begin{tabular}{l*{6}{c}}
			\toprule
			\midrule
			\textbf{Metric} &\textbf{Baseline1}&\textbf{Baseline2}&  \textbf{Proposed}\\
			\midrule
			\textsc{Rouge-1} & 0.4912 & 0.5093 & 0.5841 \\
			\textsc{Rouge-2} & 0.3050 & 0.3131 & 0.4284 \\
			\textsc{Rouge-su4} & 0.2867 & 0.3002 & 0.3950 \\
			\textsc{Bleu-4} & 0.4510 & 0.5144 & 0.6913 \\
			\midrule
			\bottomrule
		\end{tabular}
		\caption{Average scores by automatic evaluation over the normal and diverse clusters}
		\label{tbl:avg_auto_scores}
	\end{center}
\end{table}

To make the candidate and reference summaries comparable, a process of manual MWE detection is performed on the reference summaries and the MWE components are merged by three native annotators. In details, automatic evaluation packages use WordNet to compare the synonyms in candidate and reference summaries. WordNet puts hyphenation on synonyms, e.g., kick-the-bucket, so annotators hyphenate MWEs in their summaries to be used in these packages. Then, the synonym properties are set in these packages to consider the synsets. Thus, \textit{n}-words MWEs are linked to their \textit{one}-word synonyms in the candidate summary. The overall results support our hypothesis that using the POS-LM for re-ranking the compression candidates, results in more grammatical compressions, especially for diverse clusters. This issue is confirmed by 4-grams \textsc{Bleu}, which shows the grammaticality enhancement rather than the informativity. Meanwhile, we try to simultaneously improve the informativity by identifying and merging MWEs along with mapping the synonyms.

\begin{figure}[h!]
	\centering
	\includegraphics[scale=0.8]{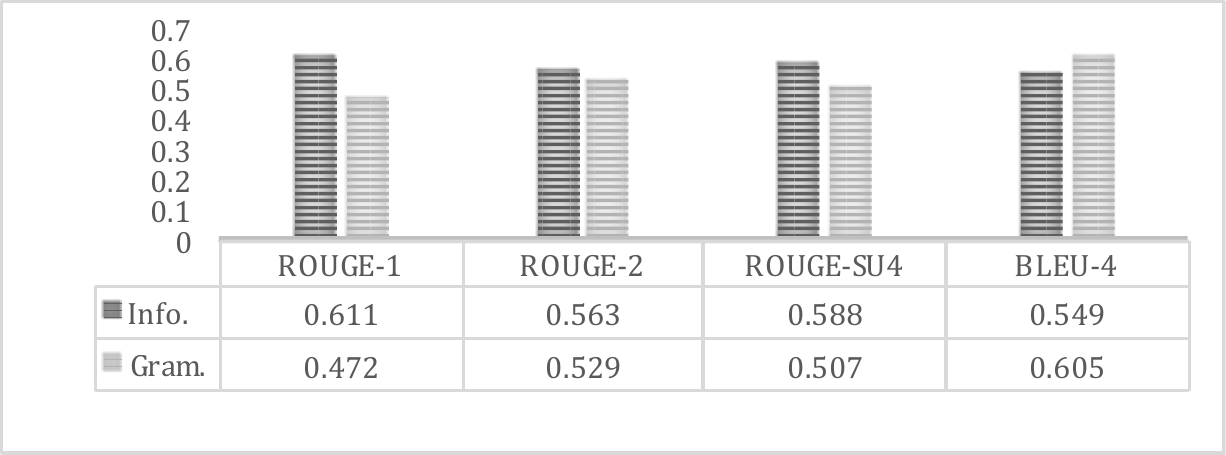}
	\caption{The effectiveness of Rouge and Bleu}
	\label{fig:eff_rouge_bleu}
\end{figure}

Furthermore, the effectiveness of \textsc{Rouge} and \textsc{Bleu} is studied using the Pearson's correlation coefficient. We found that \textsc{Rouge} shows a better correlation with informativity, while the \textsc{Bleu} correlates better with grammaticality. Overall, the results in Figure \ref{fig:eff_rouge_bleu} show high correlation (0.5 $\sim$ 1.0) between the automatic evaluation results and human ratings for both \textsc{Rouge} and \textsc{Bleu}. The main reason may be the simulation of factors that humans usually consider for summarization, such as merging and mapping strategies, along with the syntactic criteria employed by POS-LM. 

\begin{figure}[h!]
	\centering
	\includegraphics[scale=0.8]{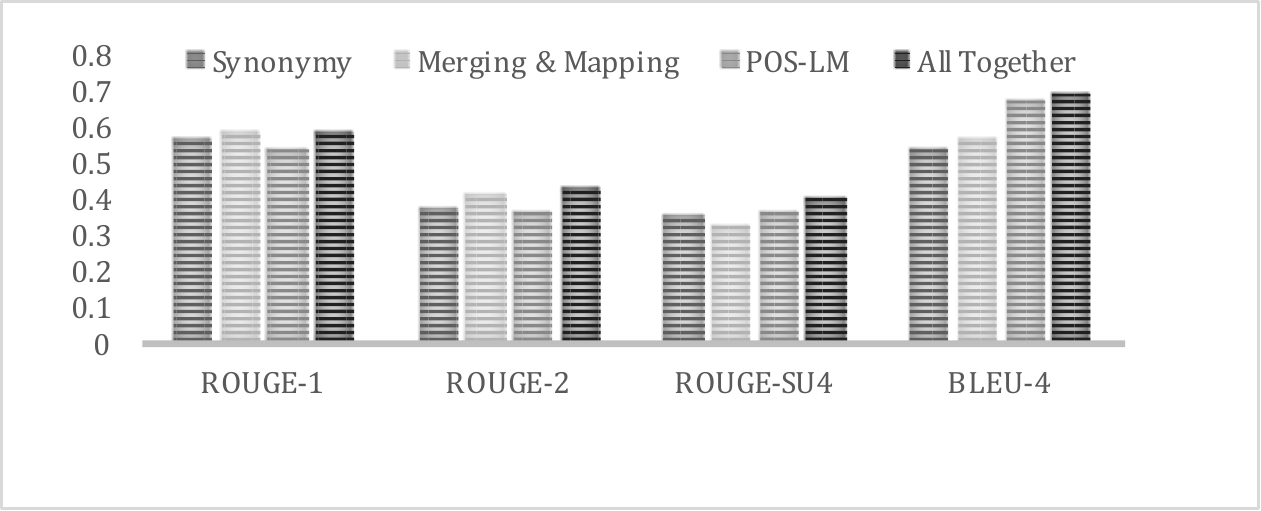}
	\caption{The effects of the improvements separately}
	\label{fig:eff_imprv_sep}
\end{figure}

To investigate the impact of each improvement separately, we have also conducted separate experiments over the prepared dataset.  The results are shown in Figure \ref{fig:eff_imprv_sep} and the related data are provided in Table \ref{tbl:eff_imprv_sep}. In our work, merging and mapping strategies significantly increase the informativity of the compressions. So, their computed scores by \textsc{Rouge} are higher than the score of POS-LM. However, the combination of MWE merging and mapping gets a slightly lower score from \textsc{Rouge-su4}. One reason may be that usage of synonymy only for MWEs and ignoring other \textit{one}-word synonym mapping causes a more diverse graph, which slightly decreases the informativity and grammaticality of compressed sentences. Meanwhile, POS-LM gets better scores from \textsc{Bleu-4}, which indicates the grammaticality enhancement rather than the informativity.  

\begin{table}[h!]

	\begin{center}	
		\begin{tabular}{l*{6}{c}}
			\toprule
			\midrule
			\textbf{Metric} &\textbf{Synonymy}&\textbf{Merg/Map}&  \textbf{POS-LM} &  \textbf{All} \\
			\midrule
			\textsc{Rouge-1} & 0.5659 & 0.5820 & 0.5381 & 0.5841 \\
			\textsc{Rouge-2} & 0.3723 & 0.4087 & 0.3599 & 0.4284 \\
			\textsc{Rouge-su4} & 0.3508 & 0.3254 & 0.3629 & 0.3950 \\
			\textsc{Bleu-4} & 0.5340 & 0.5601 & 0.6725 & 0.6913 \\
			\midrule
			\bottomrule
		\end{tabular}
		\caption{The effects of the improvements separately}
		\label{tbl:eff_imprv_sep}
	\end{center}
\end{table}

\section{Conclusions}\label{conclusions}

In a nutshell, we have presented our attempt in using MWEs, Synonymy and POS-based language modeling to tackle one of the pain points of MSC, which is improving both informativity and grammaticality at the same time. By manual and automatic (\textsc{Rouge} and \textsc{Bleu}) evaluations, experiments using a constructed English newswire dataset show that our approach outperforms the competitive baselines. In particular, the proposed merging and mapping strategies, along with the grammar-enhanced POS-LM re-ranking method, ameliorate both informativity and grammaticality of the compressions, with an improved compression ratio. 

%% bib
\bibliographystyle{plain} 
\bibliography{sample}

\begin{thebibliography}{10}

\bibitem{acosta2011identification}
Otavio~Costa Acosta, Aline Villavicencio, and Viviane~P Moreira.
\newblock Identification and treatment of multiword expressions applied to
  information retrieval.
\newblock In {\em Proceedings of the Workshop on Multiword Expressions: from
  Parsing and Generation to the Real World}, pages 101--109. Association for
  Computational Linguistics, 2011.

\bibitem{artstein2008inter}
Ron Artstein and Massimo Poesio.
\newblock Inter-coder agreement for computational linguistics.
\newblock {\em Computational Linguistics}, 34(4):555--596, 2008.

\bibitem{baldwin2010multiword}
Timothy Baldwin and Su~Nam Kim.
\newblock Multiword expressions.
\newblock {\em Handbook of Natural Language Processing, second edition. Morgan
  and Claypool}, 2010.

\bibitem{boudin2013keyphrase}
Florian Boudin and Emmanuel Morin.
\newblock Keyphrase extraction for n-best reranking in multi-sentence
  compression.
\newblock In {\em North American Chapter of the Association for Computational
  Linguistics (NAACL)}, 2013.

\bibitem{clarke2006models}
James Clarke and Mirella Lapata.
\newblock Models for sentence compression: A comparison across domains,
  training requirements and evaluation measures.
\newblock In {\em Proceedings of the 21st International Conference on
  Computational Linguistics and the 44th annual meeting of the Association for
  Computational Linguistics}, pages 377--384. Association for Computational
  Linguistics, 2006.

\bibitem{clarke2007modelling}
James Clarke and Mirella Lapata.
\newblock Modelling compression with discourse constraints.
\newblock In {\em EMNLP-CoNLL}, pages 1--11, 2007.

\bibitem{dang2005overview}
Hoa~Trang Dang.
\newblock Overview of duc 2005.
\newblock In {\em Proceedings of the document understanding conference}, pages
  1--12, 2005.

\bibitem{elsner2011learning}
Micha Elsner and Deepak Santhanam.
\newblock Learning to fuse disparate sentences.
\newblock In {\em Proceedings of the Workshop on Monolingual Text-To-Text
  Generation}, pages 54--63. Association for Computational Linguistics, 2011.

\bibitem{filippova2010multi}
Katja Filippova.
\newblock Multi-sentence compression: finding shortest paths in word graphs.
\newblock In {\em Proceedings of the 23rd International Conference on
  Computational Linguistics}, pages 322--330. Association for Computational
  Linguistics, 2010.

\bibitem{filippova2008sentence}
Katja Filippova and Michael Strube.
\newblock Sentence fusion via dependency graph compression.
\newblock In {\em Proceedings of the Conference on Empirical Methods in Natural
  Language Processing}, pages 177--185. Association for Computational
  Linguistics, 2008.

\bibitem{galley2007lexicalized}
Michel Galley and Kathleen McKeown.
\newblock Lexicalized markov grammars for sentence compression.
\newblock In {\em HLT-NAACL}, pages 180--187, 2007.

\bibitem{ganesan2010opinosis}
Kavita Ganesan, ChengXiang Zhai, and Jiawei Han.
\newblock Opinosis: a graph-based approach to abstractive summarization of
  highly redundant opinions.
\newblock In {\em Proceedings of the 23rd international conference on
  computational linguistics}, pages 340--348. Association for Computational
  Linguistics, 2010.

\bibitem{goodman2001bit}
Joshua~T Goodman.
\newblock A bit of progress in language modeling.
\newblock {\em Computer Speech \& Language}, 15(4):403--434, 2001.

\bibitem{heeman1998pos}
Peter~A Heeman.
\newblock Pos tagging versus classes in language modeling.
\newblock In {\em Proceedings of the 6th Workshop on Very Large Corpora,
  Montreal}, 1998.

\bibitem{jackendoff1997architecture}
Ray Jackendoff.
\newblock {\em The architecture of the language faculty}.
\newblock Number~28. MIT Press, 1997.

\bibitem{jing2000sentence}
Hongyan Jing.
\newblock Sentence reduction for automatic text summarization.
\newblock In {\em Proceedings of the sixth conference on Applied natural
  language processing}, pages 310--315. Association for Computational
  Linguistics, 2000.

\bibitem{koehn2008towards}
Philipp Koehn, Abhishek Arun, and Hieu Hoang.
\newblock Towards better machine translation quality for the german--english
  language pairs.
\newblock In {\em Proceedings of the Third Workshop on Statistical Machine
  Translation}, pages 139--142. Association for Computational Linguistics,
  2008.

\bibitem{koehn2007moses}
Philipp Koehn, Hieu Hoang, Alexandra Birch, Chris Callison-Burch, Marcello
  Federico, Nicola Bertoldi, Brooke Cowan, Wade Shen, Christine Moran, Richard
  Zens, et~al.
\newblock Moses: Open source toolkit for statistical machine translation.
\newblock In {\em Proceedings of the 45th annual meeting of the ACL on
  interactive poster and demonstration sessions}, pages 177--180. Association
  for Computational Linguistics, 2007.

\bibitem{kulkarni2011jmwe}
Nidhi Kulkarni and Mark~Alan Finlayson.
\newblock jmwe: A java toolkit for detecting multi-word expressions.
\newblock In {\em Proceedings of the Workshop on Multiword Expressions: from
  Parsing and Generation to the Real World}, pages 122--124. Association for
  Computational Linguistics, 2011.

\bibitem{lin2004rouge}
Chin-Yew Lin.
\newblock Rouge: A package for automatic evaluation of summaries.
\newblock In {\em Text summarization branches out: Proceedings of the ACL-04
  workshop}, volume~8, 2004.

\bibitem{mcdonald2006discriminative}
Ryan~T McDonald.
\newblock Discriminative sentence compression with soft syntactic evidence.
\newblock In {\em EACL}, 2006.

\bibitem{miller1995wordnet}
George~A Miller.
\newblock Wordnet: a lexical database for english.
\newblock {\em Communications of the ACM}, 38(11):39--41, 1995.

\bibitem{monz2011statistical}
Christof Monz.
\newblock Statistical machine translation with local language models.
\newblock In {\em Proceedings of the Conference on Empirical Methods in Natural
  Language Processing}, pages 869--879. Association for Computational
  Linguistics, 2011.

\bibitem{papineni2002bleu}
Kishore Papineni, Salim Roukos, Todd Ward, and Wei-Jing Zhu.
\newblock Bleu: a method for automatic evaluation of machine translation.
\newblock In {\em Proceedings of the 40th annual meeting on association for
  computational linguistics}, pages 311--318. Association for Computational
  Linguistics, 2002.

\bibitem{popovic2012morpheme}
Maja Popovi{\'c}.
\newblock Morpheme-and pos-based ibm1 scores and language model scores for
  translation quality estimation.
\newblock In {\em Proceedings of the Seventh Workshop on Statistical Machine
  Translation}, pages 133--137. Association for Computational Linguistics,
  2012.

\bibitem{sag2002multiword}
Ivan~A Sag, Timothy Baldwin, Francis Bond, Ann Copestake, and Dan Flickinger.
\newblock Multiword expressions: A pain in the neck for nlp.
\newblock In {\em Computational Linguistics and Intelligent Text Processing},
  pages 1--15. Springer, 2002.

\bibitem{sharifi2010experiments}
Beaux Sharifi, Mark-Anthony Hutton, and Jugal~K Kalita.
\newblock Experiments in microblog summarization.
\newblock In {\em Social Computing (SocialCom), 2010 IEEE Second International
  Conference on}, pages 49--56. IEEE, 2010.

\bibitem{stolcke2002srilm}
Andreas Stolcke et~al.
\newblock Srilm-an extensible language modeling toolkit.
\newblock In {\em INTERSPEECH}, 2002.

\bibitem{toutanova2003feature}
Kristina Toutanova, Dan Klein, Christopher~D Manning, and Yoram Singer.
\newblock Feature-rich part-of-speech tagging with a cyclic dependency network.
\newblock In {\em Proceedings of the 2003 Conference of the North American
  Chapter of the Association for Computational Linguistics on Human Language
  Technology-Volume 1}, pages 173--180. Association for Computational
  Linguistics, 2003.

\bibitem{tzouridis2014learning}
Emmanouil Tzouridis, Jamal~Abdul Nasir, LUMS Lahore, and Ulf Brefeld.
\newblock Learning to summarise related sentences.
\newblock In {\em The 25th International Conference on Computational
  Linguistics (COLING’14), Dublin, Ireland, ACL}, 2014.

\bibitem{wang2008multi}
Dingding Wang, Tao Li, Shenghuo Zhu, and Chris Ding.
\newblock Multi-document summarization via sentence-level semantic analysis and
  symmetric matrix factorization.
\newblock In {\em Proceedings of the 31st annual international ACM SIGIR
  conference on Research and development in information retrieval}, pages
  307--314. ACM, 2008.

\end{thebibliography}

\end{document}